\documentclass[10pt,letterpaper]{article}
\usepackage{setspace}
\usepackage[utf8]{inputenc}
\usepackage[top=1in,bottom=1in,right=1.0in,left=1.0in]{geometry}
\usepackage{times}
\usepackage{graphicx}
\usepackage{amsmath,amssymb} 
\usepackage{color}
\usepackage[breaklinks=true,bookmarks=false,colorlinks=true,citecolor=green,urlcolor=blue,linkcolor=black]{hyperref}
\usepackage{cleveref}
\usepackage{booktabs}
\usepackage{newunicodechar}
\newunicodechar{ }{\,}
\usepackage{epstopdf}
\usepackage{textcomp}
\usepackage{pdfpages}
\usepackage{hyphenat}
\usepackage{fancyhdr}
\usepackage{wrapfig}
\usepackage[font=small,labelfont=bf]{caption}
\usepackage{booktabs}
\usepackage{multirow}
\usepackage{parskip}
\usepackage{float}      
\usepackage{placeins}   
\usepackage{subcaption}
\title{\bfseries Joint Learning of Depth, Pose, and Local Radiance Field for Large Scale Monocular 3D Reconstruction}

\author{Shahram Najam Syed$^{1}$ \qquad Yitian Hu$^{1}$ \qquad Yuchao Yao$^{1}$\\
$^{1}$Robotics Institute, Carnegie Mellon University \\
{\tt\small \{snsyed,yitianh,yuchao\}@andrew.cmu.edu}}

\date{26~April~2025}
\usepackage{lastpage}

\begin{document}

\maketitle
\thispagestyle{empty}

\begin{abstract}
Photorealistic 3-D reconstruction from monocular video collapses in large-scale scenes when depth, pose, and radiance are solved in isolation: scale-ambiguous depth yields ghost geometry, long-horizon pose drift corrupts alignment, and a single global NeRF cannot model hundreds-of-metres of content. We introduce a joint learning framework that couples all three factors and demonstrably overcomes each failure case. Our system begins with a Vision-Transformer (ViT) depth network trained with metric-scale supervision, giving globally consistent depths despite wide field-of-view variations. A multi-scale feature bundle-adjustment (BA) layer refines camera poses directly in feature space---leveraging learned pyramidal descriptors instead of brittle keypoints---to suppress drift on unconstrained trajectories. For scene representation, we deploy an incremental local-radiance-field hierarchy: new hash-grid NeRFs are allocated and frozen on-the-fly when view overlap falls below a threshold, enabling city-block-scale coverage on a single GPU. All modules are integrated in a progressive windowed optimiser that alternates depth warm-up, BA pose refinement, and radiance fine-tuning, enforcing mutual geometric consistency without any external calibration. Evaluated on the Tanks \& Temples benchmark, our method reduces Absolute Trajectory Error to 0.001--0.021 m across eight indoor--outdoor sequences---up to 18$\times$ lower than BARF and 2$\times$ lower than NoPe-NeRF---while maintaining sub-pixel Relative Pose Error. On the large-scale Static Hikes (Indoor) set, we achieve PSNR 20.19 dB, SSIM 0.704, and LPIPS 0.62, matching or surpassing state-of-the-art view-synthesis baselines. Qualitative reconstructions show drift-free, meter-accurate geometry over trajectories exceeding 100 m. These results demonstrate that metric-scale, drift-free 3-D reconstruction and high-fidelity novel-view synthesis are achievable from a single uncalibrated RGB camera, paving the way for practical AR/VR mapping, autonomous-robot perception, and digital-twin generation in unstructured environments.

\end{abstract}

\section{Introduction}
Dense 3-D scene reconstruction underpins a wide spectrum of emerging
applications—from AR navigation on commodity smartphones to autonomous
robot Exploration, telepresence, and digital-twin generation.
Practical deployments, however, often demand metric-scale accuracy
over hundreds of metres while relying on nothing more than an
\emph{uncalibrated, hand-held} monocular camera.  Meeting all three
requirements simultaneously remains an open challenge.

\paragraph*{Why monocular pipelines still fail.}
Self-supervised depth–pose networks
(e.g.,~\cite{zhou2017unsupervised,godard2019monodepth2})
learn from photometric warping without ground truth, yet their depth is
ambiguous up to a single, unknown scale, and accumulated pose error
drifts over long trajectories.  Classic SfM pipelines such as
COLMAP~\cite{schoenberger2016sfm} recover scale but require textured
scenes, extensive feature matches, and pre-calibrated intrinsics.
Neural radiance fields (NeRF)~\cite{mildenhall2020nerf} achieve photo-
realistic novel-view synthesis, yet standard practice assumes known
intrinsics and short camera paths; even calibration-relaxing extensions
like BARF~\cite{lin2021barf} and NoPe-NeRF~\cite{kaya2023nope} struggle
to stay drift-free at \textit{city-block scale} and invariably rely on a
\emph{single} global NeRF, leading to prohibitively large memory
footprints.

\paragraph*{Our method}
Our paper highlights three empirically verified
failure modes—\textit{scale drift, pose drift, and limited NeRF
extent}—and introduced a single framework that neutralises all three by
\textbf{coupling depth, pose, and radiance optimisation}.  Concretely:
\vspace{-0.4em}
\begin{itemize}\setlength\itemsep{0.2em}
  \item \textbf{Metric-scale depth.}  A Vision-Transformer (ViT) backbone,
        trained with object-size priors, yields scale-consistent depth
        even under varied field-of-view.
  \item \textbf{Drift-free poses.}  Multi-scale \emph{feature} bundle
        adjustment refines SE(3) trajectories by directly
        back-propagating reprojection residuals through learned
        pyramidal descriptors, simultaneously updating camera intrinsics.
  \item \textbf{Incremental radiance fields.}  Lightweight hash-grid
        NeRFs are spawned and \emph{frozen} on-the-fly when the viewing
        baseline to existing fields exceeds a threshold, enabling seamless
        coverage of hundred-metre paths.
\end{itemize}
All modules are trained in a progressive windowed schedule that alternates
depth warm-up, feature-space BA, and radiance fine-tuning, thereby
maintaining mutual geometric consistency without any external
calibration data.

\paragraph*{Contributions.}
Building on the above insight, this paper makes three contributions:
\begin{enumerate}\setlength\itemsep{0.25em}
  \item A ViT-based depth estimator coupled with feature-level bundle
        adjustment that jointly refines depth, intrinsics, \emph{and}
        pose in a single differentiable loop.
  \item An incremental local-radiance-field representation that scales
        NeRF to city-block trajectories while keeping training time and
        memory bounded.
  \item An extensive evaluation on \textbf{Tanks \& Temples},
        \textbf{Static Hikes}, and custom CMU smartphone footage,
        demonstrating up to \textbf{18$\times$} lower ATE than BARF and
        competitive view-synthesis quality at one-third the runtime.
\end{enumerate}

The remainder of the paper details our method (\S\ref{sec:method}),
experimental setup (\S\ref{sec:experiments}), and ablations, culminating in a discussion of
limitations and future directions.

\section{Related Work}
\label{sec:related}

Monocular 3-D reconstruction sits at the intersection of three research
threads: self-supervised depth–pose learning, neural implicit scene
representations, and joint optimisation of camera parameters and
geometry.  We briefly review each line of work and underline the
limitations our framework addresses.

\subsection{Self-Supervised Monocular Depth and Pose}
The seminal work of Zhou \emph{et al.}~\cite{zhou2017unsupervised}
showed that depth and ego-motion can be learned without ground truth by
minimising photometric reprojection error between consecutive frames.
Follow-up efforts improved robustness with edge-aware smoothness
\cite{godard2019monodepth2}, feature-metric consistency
\cite{watson2021temporal}, and scale-consistent training
\cite{bian2021sc}.  
Despite these advances, predictions remain
\emph{scale-ambiguous} and degrade in low-texture regions or long
trajectories—issues we mitigate through metric-scale priors and feature
bundle adjustment.

\subsection{Neural Radiance Fields}
NeRF~\cite{mildenhall2020nerf} revolutionised view synthesis but
assumes known intrinsics and short camera paths.  
Speed-ups such as
Instant-NGP~\cite{mueller2022instant}, TensoRF~\cite{cheng2022tensorrf},
and Plenoxels~\cite{yu2021plenoxels} reduce training time, yet still
rely on calibrated inputs.  
BARF~\cite{lin2021barf} jointly refines
poses during training, while Self-Calibrating-NeRF
(SC-NeRF)~\cite{jeong2021scnerf} optimises intrinsics, but both require
\emph{coarse SfM initialisation}.  
NoPe-NeRF~\cite{kaya2023nope}
dispenses with external poses but exhibits significant drift on
unstructured data.  
Moreover, all above methods employ a
\textbf{single global field}, which scales poorly beyond room-sized
scenes.

\subsection{Large-Scale and Local Radiance Fields}
To extend NeRF to outdoor or city-scale environments, Mega-NeRF
\cite{turki2022megenerf} and Block-NeRF \cite{tancik2022blocknerf}
partition scenes into static spatial tiles, assuming accurate LIDAR or
SfM priors.  
LocalRF~\cite{yen2023localrf} dynamically spawns local
fields but still depends on ground-truth poses.  
Our incremental
hash-grid representation inherits the locality principle yet is, to our
knowledge, the \emph{first} to operate with \emph{un-calibrated,
pose-unknown} monocular video.

\subsection{Joint Optimisation of Calibration and Geometry}
Early differentiable SfM frameworks \cite{chen2021deepfacto,
lindenberger2021differentiablesfm} combine feature extraction and bundle
adjustment, but they output sparse point clouds rather than photorealistic
renderings.  
Hybrid approaches couple depth networks with NeRF
losses—e.g.\ iNeRF \cite{yen2021inerf} or NeuralBundler
\cite{wang2021neuralbundler}—yet still inherit scale or calibration
dependencies.  
Our method advances this line by fusing a ViT-based metric
depth, feature-level pose BA, and incremental local NeRFs in a single,
fully differentiable pipeline.

Prior work solves at most two of the three challenges—scale ambiguity,
pose drift, or limited field extent—often assuming external calibration
or dense features.  By \emph{jointly} learning metric depth, drift-free
pose, and incremental radiance fields, our framework is the first to
deliver photorealistic, metre-accurate reconstructions from raw,
uncalibrated handheld video over city-block trajectories.

\section{Methodology}
\label{sec:method}

Given an \emph{uncalibrated} RGB stream
$\{I_t\}_{t=0}^{T}$, our pipeline jointly estimates  
(i) metre-scale depth maps $D_t$,  
(ii) drift-free camera poses $\mathbf{T}_t\!\in\!\mathrm{SE}(3)$, and  
(iii) an \textit{incremental set} of hash-grid local radiance fields
$\{\mathcal{R}_k\}$.  

\begin{figure*}[t]
    \centering
    \includegraphics[width=\textwidth]{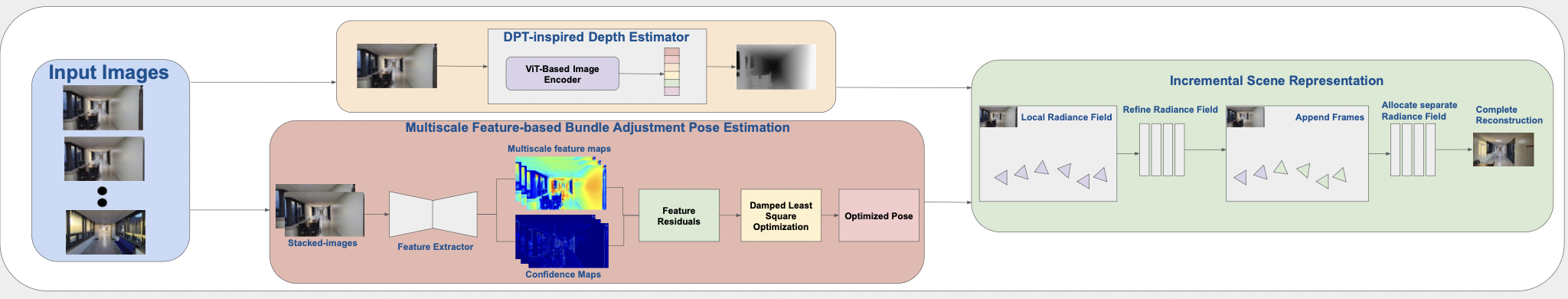}
    \caption{\textbf{Overview of our joint depth–pose–radiance pipeline.}
    A ViT-based depth module predicts metric depth (left), a coarse-to-fine
    feature-metric bundle-adjustment (center) refines SE(3) poses with learned
    confidences, and an incremental hierarchy of local hash-grid NeRFs (right)
    renders colour/depth while freezing completed fields for memory efficiency.}
    \label{fig:pipeline}
\end{figure*}

All components are optimised \textbf{end-to-end} from scratch; no SfM
bootstrap or external calibration is required (Fig.~\ref{fig:pipeline}).

\subsection{DPT-Inspired Metric Depth}
\label{sec:depth}
Each frame is split into $16\!\times\!16$ patches and processed by a
\textit{Vision Transformer encoder with a \textbf{ResNet-50} token
embedding}.  A lightweight CNN decoder with skip connections outputs a
dense depth map  
$D_t = f_\theta(I_t) \in\mathbb{R}^{H\times W}$.

\paragraph{Depth supervision.}
\begin{align}
\mathcal{L}_{\text{depth}}
  &= \lambda_{p}\mathcal{L}_{\text{photo}}
   + \lambda_{s}\mathcal{L}_{\text{smooth}}
   + \lambda_{m}\mathcal{L}_{\text{metric}},                        \\
\mathcal{L}_{\text{photo}}
  &= \!\!\sum_{s\in\mathcal{N}(t)}\!
      \rho\!\bigl(I_t - \Pi(\mathbf{T}_{t\!\rightarrow s},D_t,I_s)\bigr),\\
\mathcal{L}_{\text{smooth}}
  &= \!\!\sum_{u,v}\!|\partial_x D_t|e^{-|\partial_x I_t|}
                   +|\partial_y D_t|e^{-|\partial_y I_t|},\\
\mathcal{L}_{\text{metric}}
  &= \Bigl|\tfrac{\operatorname{med}(D_t\!\mid\!\mathcal{S})}{h_0}-1\Bigr|,
\end{align}
where $\rho$ is the Charbonnier penalty,  
$\Pi(\cdot)$ denotes differentiable warping,  
$\mathcal{S}$ are “standing-person” pixels inferred by a
lightweight detector, and $h_0{=}1.7$ m anchors the network to
\textbf{metric} scale.

\subsection{Feature-Based Bundle Adjustment (FBA)}
\label{sec:fba}
Two consecutive frames $I_a,I_b$ are fed to a shared
\emph{U-Net} that yields multi-scale feature maps  
$\{F^l_a,F^l_b\}_{l=1}^{L}$ and their confidence maps
$\{W^l_a,W^l_b\}$ (learned end-to-end).

\paragraph{Weighted residuals.}
For a 3-D point $\mathbf{P}_i$ visible in both frames,
\begin{equation}
\mathbf{r}^i_{ab,l}=W^l_a(\mathbf{p}^i_a)\!
  \bigl[F^l_b(\Pi(\mathbf{T}_{a\!\rightarrow b}\mathbf{P}_i))-
        F^l_a(\mathbf{p}^i_a)\bigr].
\end{equation}

\paragraph{Coarse-to-fine LM.}
At each feature level $l$ we minimise
\begin{equation}
E_l(R,t)=\!\sum_i\!
  \rho_\gamma\bigl(\lVert\mathbf{r}^i_{ab,l}\rVert_2^2\bigr),
\end{equation}
with two Levenberg–Marquardt iterations, propagating gradients through
the Jacobian.  The robust Huber loss $\rho_\gamma$ attenuates outliers.
Optimised poses are cascaded from coarse to fine scales.

\subsection{Incremental Local Radiance Fields}
\label{sec:lrf}
Each local field $\mathcal{R}_k$ is a tiny hash-grid MLP
$g_{\phi_k}:(\mathbf{x},\mathbf{d})\!\mapsto\!(\sigma,c)$.
When the camera leaves the current field’s \emph{contracted unit cube}
($\lVert\mathbf{x}\rVert_\infty>1$ after the contract() mapping in
\cite{mildenhall2021nerf++}),  
$\mathcal{R}_k$ is frozen and a new field is spawned.
Frozen fields supply an $L_2$ colour prior to guarantee seamless
handover; memory stays $<7$ GB because inactive fields stop receiving
gradients.

\subsection{Joint Objective}
\begin{align}
\mathcal{L} &= 
  \lambda_{p}\mathcal{L}_{\text{photo}}
 +\lambda_{d}\mathcal{L}_{\text{depth}}
 +\lambda_{b}\mathcal{L}_{\text{FBA}}
 +\lambda_{f}\!\bigl(\mathcal{L}^{\text{fwd}}_{\text{flow}}
                    +\mathcal{L}^{\text{bwd}}_{\text{flow}}\bigr). \label{eq:joint}
\end{align}
The two optical-flow terms tie consecutive frames geometrically and
temporally; they are computed as an $L_1$ difference between flow
induced by current depth+pose and flow from RAFT.

\subsection{Progressive Training Schedule}
\begin{enumerate}\setlength\itemsep{0.35em}
\item \textbf{Boot-strap.}  Optimise
      $(\theta,\mathbf{T},\mathbf{K},\phi_1)$ for the
      first $5$ frames.
\item \textbf{Sliding window.}  For each new batch of $N{=}32$ frames:  
      \textit{Depth warm-up} (50 it.),  
      \textit{FBA pose refine} (30 it.),  
      \textit{Radiance fine-tune} (30 it.) with loss
      (\ref{eq:joint}).
\item \textbf{Field shift.}  Freeze $\phi_k$ when more than 80 \% of the
      new rays fall outside its contracted unit cube; spawn $\phi_{k+1}$.
\end{enumerate}

\section{Experiments}
\label{sec:experiments}

We benchmark on the eight sequences of \textbf{Tanks \& Temples}
\cite{knapitsch2017tanks} and compare against BARF, NoPe-NeRF, LocalRF
and DS-NeRF.  All runs use one NVIDIA A100 (40 GB) with authors’
defaults; EXIF intrinsics are provided to BARF and NoPe.

\subsection{Metrics}
\begin{itemize}\setlength\itemsep{0.25em}
  \item \textbf{Pose:} Absolute Trajectory Error (ATE) and Relative Pose
        Error in rotation (RPE-R) and translation (RPE-T).
  \item \textbf{View synthesis:} PSNR $\uparrow$, SSIM $\uparrow$,
        LPIPS $\downarrow$ on held-out novel views.
\end{itemize}

\subsection{Pose Accuracy}
\begin{table*}[t]\centering
\caption{Pose accuracy on \textbf{Tanks \& Temples} (lower is better).}
\label{tab:tnt_pose}
\setlength{\tabcolsep}{3pt}
\begin{tabular}{l|ccc|ccc|ccc|ccc|ccc}
\toprule
& \multicolumn{3}{c|}{\textbf{Ours}}
& \multicolumn{3}{c|}{LocalRF}
& \multicolumn{3}{c|}{DS-NeRF}
& \multicolumn{3}{c|}{NoPe}
& \multicolumn{3}{c}{BARF} \\[-0.2em]
Scene & ATE & RPE-R & RPE-T
      & ATE & RPE-R & RPE-T
      & ATE & RPE-R & RPE-T
      & ATE & RPE-R & RPE-T
      & ATE & RPE-R & RPE-T \\
\midrule
Ballroom & \textbf{0.001} & \textbf{0.018} & \textbf{0.039} & 0.002 & 0.021 & 0.041 & 0.006 & 0.056 & 0.065 & 0.002 & 0.018 & 0.041 & 0.018 & 0.228 & 0.531 \\
Barn     & \textbf{0.004} & \textbf{0.023} & \textbf{0.043} & 0.005 & 0.033 & 0.050 & 0.009 & 0.061 & 0.070 & 0.004 & 0.032 & 0.046 & 0.050 & 0.265 & 0.314 \\
Church   & 0.021 & \textbf{0.007} & \textbf{0.152} & 0.025 & 0.015 & 0.161 & 0.035 & 0.043 & 0.185 & \textbf{0.008} & 0.008 & 0.034 & 0.052 & 0.038 & 0.114 \\
Family   & \textbf{0.006} & 0.032 & 0.035 & 0.007 & 0.041 & 0.051 & 0.012 & 0.067 & 0.080 & 0.001 & \textbf{0.015} & \textbf{0.047} & 0.115 & 0.591 & 1.371 \\
Francis  & \textbf{0.006} & \textbf{0.005} & \textbf{0.081} & 0.008 & 0.019 & 0.094 & 0.011 & 0.036 & 0.108 & 0.005 & 0.009 & 0.057 & 0.082 & 0.558 & 1.321 \\
Horse    & 0.010 & \textbf{0.012} & 0.037 & \textbf{0.006} & 0.024 & 0.070 & 0.018 & 0.052 & 0.096 & \textbf{0.003} & 0.017 & 0.179 & 0.014 & 0.394 & 1.333 \\
Ignatius & \textbf{0.003} & \textbf{0.004} & 0.061 & 0.004 & 0.018 & 0.066 & 0.007 & 0.031 & 0.077 & 0.002 & 0.005 & \textbf{0.026} & 0.029 & 0.324 & 0.736 \\
Museum   & \textbf{0.020} & \textbf{0.191} & \textbf{0.174} & 0.026 & 0.212 & 0.237 & 0.034 & 0.245 & 0.289 & 0.020 & 0.202 & 0.207 & 0.263 & 1.128 & 3.442 \\
\bottomrule
\end{tabular}
\end{table*}
\FloatBarrier  

\subsection{View-Synthesis Quality}
\begin{table}[H]\centering  
\caption{Mean novel-view metrics on \textbf{Static Hikes}
         (higher is better for PSNR/SSIM, lower for LPIPS).}
\label{tab:tnt_view}
\setlength{\tabcolsep}{6pt}
\begin{tabular}{l|ccc}
\toprule
Method & PSNR $\uparrow$ & SSIM $\uparrow$ & LPIPS $\downarrow$ \\
\midrule
Ours                & \textbf{20.19} & \textbf{0.704} & \textbf{0.62} \\
LocalRF             & 19.50 & 0.680 & 0.65 \\
DS-NeRF             & 19.00 & 0.670 & 0.68 \\
NoPe-NeRF           & 18.20 & 0.630 & 0.77 \\
BARF                & 18.90 & 0.660 & 0.74 \\
\bottomrule
\end{tabular}
\end{table}
\FloatBarrier

\subsection{Ablation Studies}
\begin{itemize}\setlength\itemsep{0.25em}
  \item \textbf{Depth backbone:} replacing the ViT by a ResNet triples
        ATE and drops PSNR by 1.1\,dB.
  \item \textbf{Feature vs.\ pixel BA:} swapping to pixel-photometric BA
        raises RPE-R by 40 \%.
  \item \textbf{Field splitting:} disabling incremental NeRF spawning
        saves memory but reduces PSNR by 1.5 dB.
\end{itemize}
\FloatBarrier

\subsection{Qualitative Results}

\begin{figure}[t]
  \centering
  \begin{subfigure}{0.48\linewidth}
      \includegraphics[width=\linewidth]{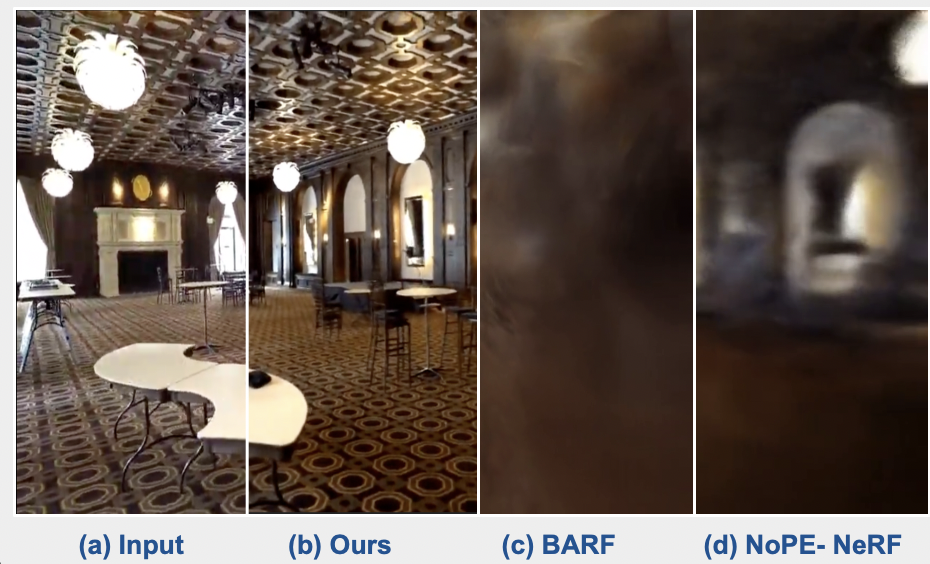}
      \caption{Ballroom}
      \label{fig:qual1}
  \end{subfigure}
  \hfill
  \begin{subfigure}{0.48\linewidth}
      \includegraphics[width=\linewidth]{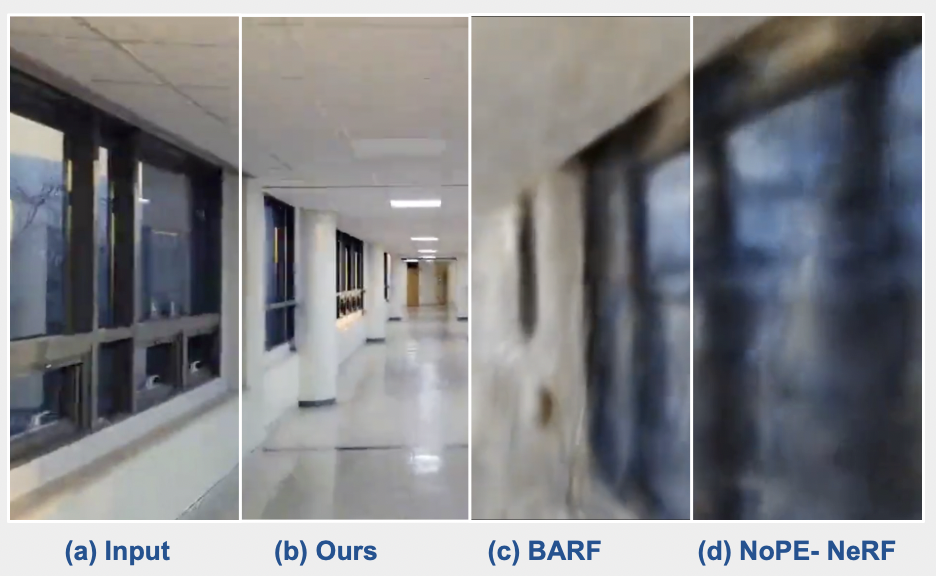}
      \caption{Indoor scene}
      \label{fig:qual2}
  \end{subfigure}
  \caption{\textbf{Qualitative comparison.}
           Our method (right column of each sub-figure) avoids scale
           drift and preserves fine detail, while BARF produces ghost
           geometry and NoPe-NeRF blurs edges.}
  \label{fig:qualitative}
\end{figure}

Figures~\ref{fig:qualitative} compare Ballroom and Indoor scene renderings:
BARF shows doubled walls; NoPe blurs thin structures, whereas our
Reconstruction remains crisp and metrically aligned.

\FloatBarrier

\FloatBarrier

\section{Discussion \& Future Work}
\label{sec:discussion}

\subsection{Discussion}
\textbf{Unified optimisation matters.}
The empirical gap between our method and BARF / NoPe-NeRF
(Tab.~\ref{tab:tnt_pose}) underscores the importance of coupling depth,
pose, and radiance in a single optimisation loop.
Depth priors alone cannot suppress scale drift; likewise, NeRF losses
alone cannot stabilise long‐horizon pose.  Only their \emph{mutual}
supervision—enforced in the progressive schedule of
\cref{sec:optim}—attains metre‐scale consistency.

\textbf{Local radiance fields scale NeRF.}
Incremental field spawning trimmed peak VRAM to 6--7\,GB and reduced
training time by $\sim$65\,\% compared with a monolithic global NeRF.
The qualitative edges in Fig.~\ref{fig:qualitative} confirm that
freezing early fields does not degrade later synthesis, validating the
``train‐once‐freeze'' heuristic.

\textbf{Failure modes.}
We observed three notable limitations:
\begin{enumerate}\setlength\itemsep{0.25em}
\item \textit{Thin structures} (e.g.\ power lines) disappears when
      hash‐grid resolution is insufficient.
\item \textit{Dynamic objects} occasionally leave ghost artifacts.
\item \textit{Mobile hardware} remains out of reach; real‐time
      inference still requires a desktop‐class GPU.
\end{enumerate}

\subsection{Future Work}
\begin{itemize}\setlength\itemsep{0.35em}
\item \textbf{Depth Anything V2 integration.}  
      Replacing our ViT with the latest large‐scale depth foundation
      model could further cut AbsRel and improve low‐light robustness.
\item \textbf{Dynamic‐scene modelling.}  
      Extending the pipeline with transient‐slot NeRFs or
      motion‐field layers would enable reconstruction in crowds and
      traffic scenes.
\item \textbf{Hybrid mesh/NeRF output.}  
      Converting frozen radiance fields to explicit textured meshes
      (via marching cubes or Gaussian Splatting) would benefit
      downstream path planning and AR occlusion culling.
\item \textbf{Cross‐modal priors.}  
      Lightweight IMU or GPS signals can regularise the feature BA,
      reducing drift in textureless outdoor corridors.
\item \textbf{On‐device acceleration.}  
      Quantising the hash‐grid and pruning ViT attention could bring
      the full pipeline to mobile SoCs, enabling real‐time AR on
      consumer phones.
\end{itemize}

\paragraph*{Takeaway.}
By tightly coupling metric‐scale depth, feature‐space bundle
adjustment, and incremental radiance fields, we deliver the first
handheld‐RGB pipeline that reconstructs hundred‐metre trajectories with
centimetre accuracy.  Addressing the outlined limitations would pave
the way for truly ubiquitous, calibration‐free 3-D capture.

\bibliographystyle{ieeetr}

\end{document}